\documentclass[10pt,twocolumn,letterpaper]{article}

\usepackage{wacv}                       

\usepackage{amsmath}
\usepackage{amssymb}
\usepackage{array}
\usepackage{siunitx}
\usepackage{pifont}
\usepackage{multirow}
\usepackage{algorithm}
\usepackage{algpseudocode}
\usepackage{tabularx}
\usepackage{booktabs}
\usepackage{graphicx}
\usepackage[protrusion=true,expansion=true]{microtype}
\usepackage{textcomp}
\usepackage[table]{xcolor}
\usepackage{placeins}
\usepackage{flafter}
\usepackage{float}
\usepackage{url}
\usepackage[T1]{fontenc}
\usepackage{enumitem}
\usepackage{algorithm}
\usepackage{amsmath}
\usepackage{amsfonts}
\usepackage{amssymb}
\usepackage{booktabs}
\usepackage{graphicx}
\usepackage{array}
\usepackage{tabularx}
\usepackage{xcolor}
\definecolor{AlgText}{HTML}{1F2937}
\definecolor{AlgGray}{HTML}{64748B}
\definecolor{AlgViolet}{HTML}{5A3FE0}
\definecolor{AlgCyan}{HTML}{1C839A}
\definecolor{AlgOrange}{HTML}{C2641A}
\definecolor{AlgBorder}{HTML}{D5DCE8}
\definecolor{AlgBg}{HTML}{FAFBFD}
\definecolor{yologroupbg}{RGB}{224, 242, 255}

\usepackage{colortbl}
\usepackage{bm}
\definecolor{bestred}{HTML}{C0392B}
\definecolor{secblue}{HTML}{2457D6}
\definecolor{upgreen}{HTML}{1A7F37}
\definecolor{mgshade}{RGB}{224, 242, 255}
\setlist[enumerate]{
  label=\arabic*),
  leftmargin=2em,
  labelwidth=0em,
  labelsep=0.5em,
  itemindent=0pt,
  align=left,
  itemsep=0.5ex,
  parsep=0pt
}

\usepackage{pifont}
\newcommand{\cmark}{\ding{51}}

\definecolor{wacvblue}{rgb}{0.21,0.49,0.74}
\usepackage[pagebackref,breaklinks,colorlinks,allcolors=wacvblue]{hyperref}

\def\wacvPaperID{1244} 
\def\confName{WACV}
\def\confYear{2027}

\title{
\textcolor[RGB]{10,75,145}{M}%
\textcolor[RGB]{45,125,190}{G}%
\textcolor[RGB]{85,170,220}{D}%
\textcolor[RGB]{65,150,210}{F}%
\textcolor[RGB]{30,105,175}{I}%
\textcolor[RGB]{0,55,130}{S}: Multi-scale Global-detail Feature Integration Strategy for Small Object Detection
}

\author{
Yuxiang Wang\textsuperscript{1}\thanks{Equal contribution.}
\quad
Xuecheng Bai\textsuperscript{2}\footnotemark[1]
\quad
Chuanzhi Xu\textsuperscript{1}\thanks{Corresponding author: chuanzhi.xu@sydney.edu.au}
\quad
Ying Zhou\textsuperscript{1}
\quad
Weidong Cai\textsuperscript{1}
\\
\textsuperscript{1}School of Computer Science, The University of Sydney, Sydney, Australia
\\
\textsuperscript{2}Aviation Traffic Control Technology (SHENZHEN) Co., Ltd., Shenzhen, China
}

\begin{document}
\maketitle
\begin{abstract}
Small-object detection in Unmanned Aerial Vehicle (UAV) imagery requires preserving weak local evidence while using broader context to separate tiny foreground targets from cluttered backgrounds. Existing multi-scale fusion methods improve feature aggregation, but they often add computation or blur fine details during repeated cross-scale fusion. The central challenge is to balance low-SNR target preservation, clutter suppression, and efficient cross-scale context exchange. To address this challenge, we propose the Multi-scale Global-detail Feature Integration Strategy (MGDFIS), a neck-level feature-fusion strategy that couples global context exchange, local-detail recovery, and pixel-level foreground-background recalibration. MGDFIS integrates three coordinated modules: FusionLock-TSS Attention for stabilizing spectral-spatial responses, Global-detail Integration for combining long-range mixing with local detail capture, and Dynamic Pixel Attention for reweighting compact foreground regions. On the controlled VisDrone setting, YOLO26m + MGDFIS improves $AP_{50:95}$ from 25.7 to 30.2 and $AP_{50}$ from 37.2 to 44.2 over the YOLO26m baseline, with 96.1 GFLOPs. Additional dataset-specific evaluations report 38.9 $AP_{50}$ and 21.9 $AP_{50:95}$ on UAVDT and 97.4 $AP_{50}$ on CARPK. The code is available at: \href{https://github.com/Bai-Xuecheng/MGDFIS}{https://github.com/Bai-Xuecheng/MGDFIS}.

\end{abstract}
\section{Introduction}

\begin{figure}[t]
    \centering
    \includegraphics[trim=6 9 3 3, clip, width=\linewidth]{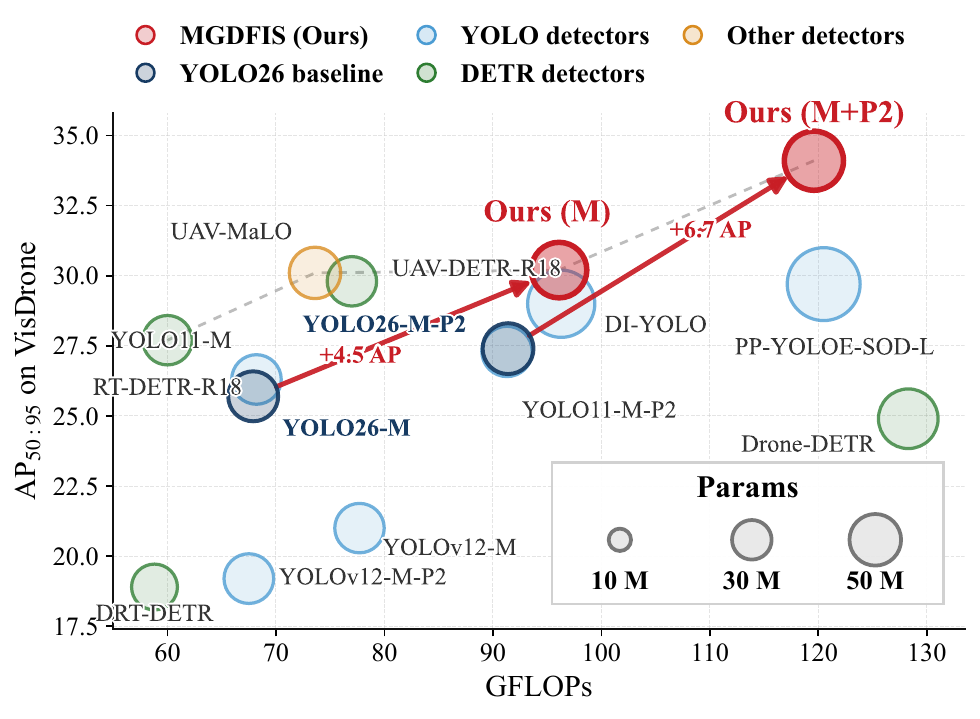}
    \caption{\textbf{Accuracy--computation trade-off on VisDrone.} The horizontal axis reports GFLOPs, the vertical axis reports $AP_{50:95}$, and bubble area denotes the number of parameters. Red arrows show the controlled gains from YOLO26 baselines to MGDFIS-enhanced variants.}
    \label{fig:visdrone_tradeoff}
\end{figure}

With the rapid development of advanced air mobility, unmanned aerial vehicles (UAVs) have become increasingly important for traffic monitoring, logistics inspection, and environmental sensing. Unlike generic object detection, UAV detection often requires fast localization of extremely small targets, typically smaller than $32 \times 32$ pixels, in real scenes with dense objects and cluttered backgrounds.

Small-object detection faces a structural tension: detectors must preserve weak local semantics while using sufficient context to distinguish true targets from textured backgrounds~\cite{li2023lsknet,xu2022nwd,ding2022dota}. The former relies on high-resolution and fine-grained features~\cite{yang2022querydet,xu2022nwd,Shi_2025_AAAI}, but repeated downsampling quickly weakens the localization evidence of small objects. The latter requires larger receptive fields and cross-region interactions~\cite{li2023lsknet,zhu2023biformer,xia2022dat}, while illumination changes and occlusion in complex scenes further reduce the foreground signal-to-noise ratio and increase computational cost. Effective small-object detection therefore requires a compact and coordinated organization of local details, global context, and foreground-background selection.

Existing methods mitigate this issue from different perspectives, including multi-scale feature fusion~\cite{zhao2024detrs,du2025cross,Shi_2025_AAAI}, high-resolution query mechanisms~\cite{yang2022querydet}, and noise-aware feature pyramids~\cite{Fang_2025_ICME}. However, they usually emphasize a single aspect of the problem. This motivates the question studied in this paper: how can local details, global context, and pixel-level foreground-background selection be unified within the detection neck?

To address this question, in this paper, we propose the \textbf{\underline{M}ulti-scale \underline{G}lobal-\underline{d}etail \underline{F}eature \underline{I}ntegration \underline{S}trategy (\underline{MGDFIS})}, a neck-replacement feature fusion design for small-object detection. MGDFIS treats fusion as a coordinated feature organization process: it stabilizes low-SNR responses before fusion, enables long-range pixels to exchange context through global-detail integration, and uses pixel-level attention to recalibrate foreground and background responses. This design makes feature interaction more selective while preserving the local shape and texture cues required by tiny targets.

We evaluate MGDFIS on VisDrone~\cite{du2019visdrone}, UAVDT, and CARPK. The results show consistent improvements across datasets, with the VisDrone accuracy--computation trade-off shown in Fig.~\ref{fig:visdrone_tradeoff}.

The main contributions are summarized as follows:

\begin{itemize}
[
    left=0pt,
    labelsep=1em,
    align=parleft,
    topsep=0pt,      
    partopsep=0pt,   
    parsep=0pt,      
    itemsep=0em    
]
  \item We propose \textbf{MGDFIS}, a neck-level feature fusion strategy for UAV small-object detection.
  \item We introduce \textbf{\underline{F}usionLock-\underline{TSS} \underline{A}ttention (\underline{FTSSA})}, a detail-processing module that stabilizes low-SNR small-object features and enhances spectral-spatial cues.
  \item We design the \textbf{\underline{G}lobal-\underline{d}etail \underline{I}ntegration (\underline{GDIM})} and \textbf{\underline{D}ynamic \underline{P}ixel \underline{A}ttention (\underline{DPAM})} for context exchange, local detail preservation, and foreground-background pixel calibration.
  \item On VisDrone, integrating MGDFIS into YOLO26m improves AP/AP50 from 25.7/37.2 to 30.2/44.2, corresponding to relative gains of 17.5\%/18.8\%. MGDFIS also shows strong performance on UAVDT and CARPK.
\end{itemize}

\begin{figure*}[t]
  \centering
  \includegraphics[trim=25 19 18 20, clip, width=\textwidth]{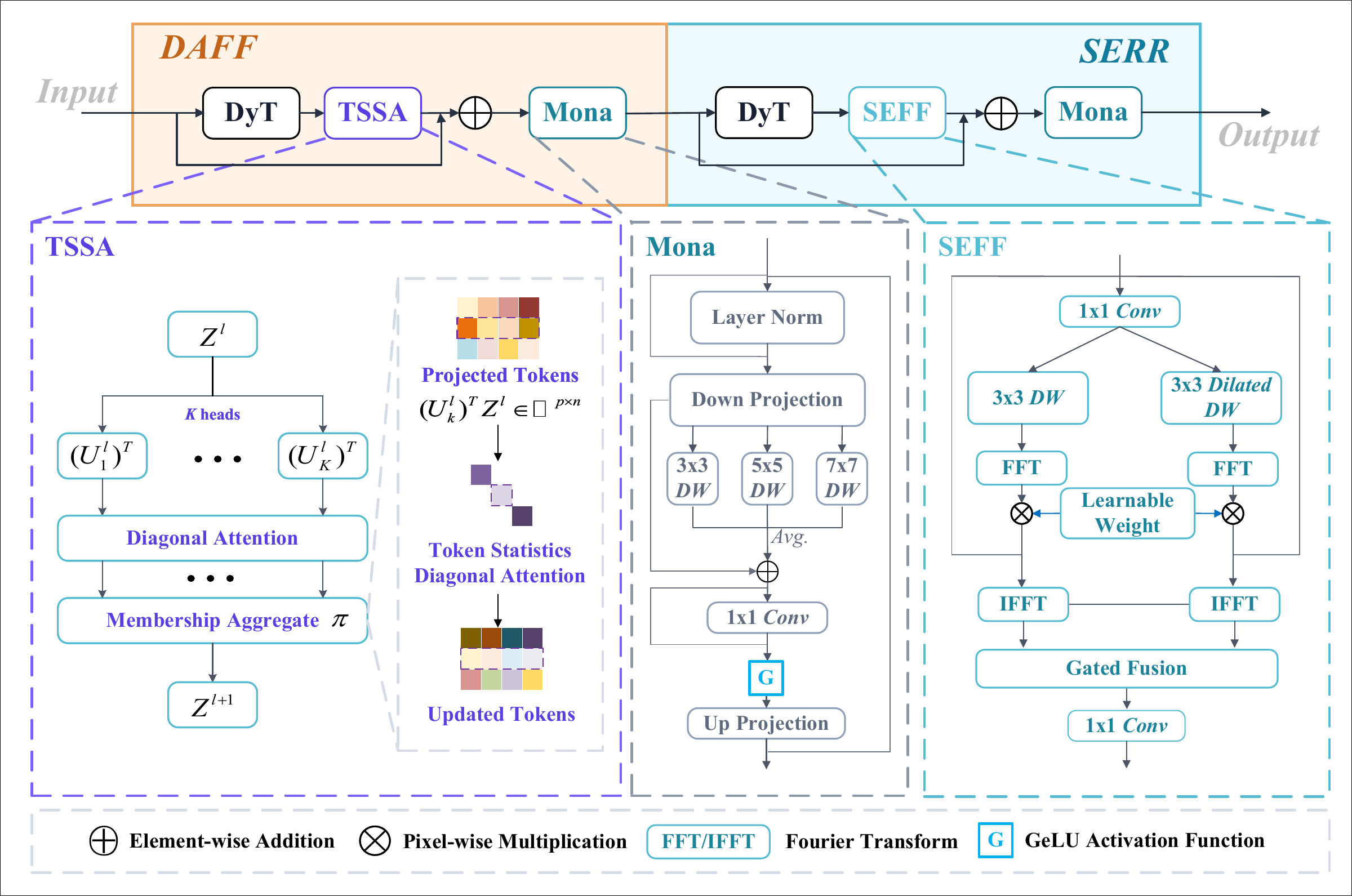}
  \caption{\textbf{FusionLock-TSS Attention} stabilizes weak small-object responses before global-detail fusion. DAFF combines DynamicTanh normalization, TSSA attention, and MonaOp spatial mixing to filter redundant responses. SERR then applies spectral enhancement and residual MonaOp refinement to produce a higher signal-to-noise feature representation.}
  \label{fig:fusionlock_tss_attention}
\end{figure*}

\section{Related Work}
This section reviews prior work related to UAV small-object detection, high-resolution detector design, and feature fusion. For a broader overview, recent surveys summarize current progress and challenges in small-object detection~\cite{wei2024review}.
\subsection{Small-Object Detection}
General object detection has rapidly evolved from CNN-based detectors~\cite{ren2015faster,lin2017focal} to end-to-end DETR-style frameworks~\cite{zhu2021deformable,zhang2023dino,zhao2024detrs}, and further to open-vocabulary vision-language detectors~\cite{li2022glip,liu2024groundingdino}. These advances, however, are largely optimized for objects with sufficient spatial support in standard benchmarks and do not transfer directly to tiny targets in UAV imagery, where instances may occupy only a few pixels~\cite{wei2024review,gong2021effective}. From a top-down viewpoint, such targets provide scarce appearance evidence; this evidence is further diluted by progressive backbone downsampling and is often scattered across large backgrounds in sparse, cluttered, or clustered layouts.

Existing methods address this challenge from two main perspectives. The first allocates high-resolution computation to more informative regions. QueryDet uses coarse low-resolution queries to guide sparse high-resolution reasoning~\cite{yang2022querydet}, while density-guided and global-context-driven aerial detectors exploit clustered object distributions or scene context to narrow the effective search space~\cite{yang2019clustered,deng2021glsan,du2023adaptive}. The second strengthens weakened target responses. Some methods recover weak features through denoising and transformer interaction, as represented by Transformer R-CNN with DeNoising FPN~\cite{liu2024dntr}; others recast UAV detection into an end-to-end DETR paradigm or redesign the feature pyramid for aerial small objects~\cite{zhang2025uav,kong2024drone,du2025cross}.

Together, these studies indicate that UAV small-object detection depends not only on stronger features, but also on where computation is allocated and how weak evidence is preserved.
\vspace{-2pt}
\subsection{Feature Enhancement and Fusion Methods}
Feature enhancement and fusion have shifted from plain cross-level aggregation toward more structured feature organization, with efforts concentrated along three axes: cross-scale information exchange, local-detail preservation, and weak-response enhancement~\cite{wang2023goldyolo,zhao2024detrs,du2025cross,Bacea_2025_WACV,zhang2025uav,Bian_2025_CVPR}. Gold-YOLO introduces a gather-and-distribute mechanism to reorganize multi-scale information flow~\cite{wang2023goldyolo}; RT-DETR decouples intra-scale interaction from cross-scale fusion in its hybrid encoder~\cite{zhao2024detrs}; and CFPT redesigns the feature pyramid with cross-layer channel and spatial attention for aerial small-object detection~\cite{du2025cross}. Complementarily, ECF-YOLOv7-Tiny improves lightweight fusion and receptive-field modeling~\cite{Bacea_2025_WACV}, UAV-DETR augments multi-scale fusion with frequency enhancement and frequency-focused downsampling to preserve boundary and texture cues~\cite{zhang2025uav}, and recent tiny-object work highlights downsampling-weakened regions through pixel-information and positional Gaussian maps~\cite{Bian_2025_CVPR}.

These methods are effective, but most still treat cross-scale exchange, detail preservation, and weak-response enhancement as separate add-ons. In contrast, MGDFIS casts the neck as a unified, evidence-aware process of organizing small-object features: within a single fusion path, it couples global--local interaction with pixel-level foreground--background calibration, thereby reducing the information dilution caused by repeated aggregation.

\section{Methodology}
MGDFIS treats UAV small-object fusion as a feature organization problem. The design follows three principles: stabilize weak responses before fusion, exchange long-range context without discarding local detail, and recalibrate compact foreground regions at the pixel level. The novelty of MGDFIS lies in the fusion strategy and in the GMM, DMM, and DPAM components, while FTSSA organizes existing attention and refinement operators into a reusable stabilization stage. GDIM then performs global-detail integration and reuses FTSSA internally for detail refinement; DPAM finally applies pixel-level foreground-background recalibration, whose benefit is strongest when high-resolution features provide sufficiently precise spatial evidence. The following sections describe the modules in this order.

\subsection{FusionLock-TSS Attention Module}
FTSSA stabilizes weak small-object evidence before global-detail fusion. Rather than using attention only as a spatial weighting step, it organizes four proven, off-the-shelf components along complementary axes: channel activation control, token-statistical dependency modeling, spatial mixing, and frequency-domain refinement. Concretely, FTSSA integrates DynamicTanh (DyT) normalization~\cite{zhu2025transformers}, Token Statistics Self-Attention (TSSA)~\cite{wu2024token}, the Multi-cognitive Visual Adapter (Mona)~\cite{yin2023adapter}, and the Spectral Enhanced Feed-Forward (SEFF) module~\cite{sun2024hybrid}, as shown in Fig.~\ref{fig:fusionlock_tss_attention}. We adopt these components without modifying their internal designs; our contribution is their organization into a single pre-fusion stabilization operator tailored to low-SNR aerial targets.
Each component handles a different source of instability. DyT regulates channel activation ranges so that low-contrast targets are less likely to be suppressed during normalization. TSSA uses token statistics to capture long-range relations with linear complexity. Mona injects multi-kernel local mixing to recover boundary and texture cues. SEFF refines the stabilized representation in the frequency domain before the residual path returns it to GDIM. FTSSA therefore acts as a pre-fusion locking stage for weak foreground evidence.
\vspace{-3pt}
\subsubsection{Dynamic Attention and Feature Fusion.}
The dynamic-attention stage implements the channel-statistical-spatial path described above. It reduces redundant channel responses, models long-range token relations, and recovers local spatial evidence before the feature enters GDIM. DyT first adapts the channel activation range, TSSA then filters statistical interference, and Mona aggregates multi-scale spatial evidence so that weak cues remain available to downstream layers.
Let the input feature be $x \in \mathbb{R}^{C \times H \times W}$. DyT introduces a learnable scalar parameter $\alpha$ and applies a tangent function for scaling and shifting, producing the refined feature $F \in \mathbb{R}^{C \times H \times W}$ as shown in Eq.~\eqref{eq:dyt}:
\begin{align}
\label{eq:dyt}
F &= \operatorname{DyT}(x) = \gamma \odot \tanh(\alpha x) + \beta
\end{align}
Next, the refined feature $F$ is fed into TSSA~\cite{wu2024token} as a token sequence. Instead of computing pairwise token affinities, TSSA derives its attention from second-order token statistics, which yields linear $O(N)$ complexity and is robust to the low signal-to-noise responses typical of small targets. The sequence is projected by a single linear layer and reweighted by per-group second-order moments, producing a group-compressed representation $F_{Com}\in \mathbb{R}^{C \times H \times W}$ as shown in Eq.~\eqref{eq:tssa}. We adopt TSSA without modification and refer readers to~\cite{wu2024token} for the full derivation.
\begin{equation}
\label{eq:tssa}
F_{Com}=\operatorname{TSSA}(F)
\end{equation}
The compressed feature $F_{Com}$ is then refined by Mona~\cite{yin2023adapter}, an adapter that combines two complementary branches: a globally scaled normalization branch and a multi-scale convolutional branch (MonaOp) that aggregates $3{\times}3$, $5{\times}5$, and $7{\times}7$ responses to recover boundary and texture cues. We likewise adopt Mona without modification and denote its output $\operatorname{Mona}(\cdot)$, which yields the dynamic-attention output defined below.

Finally, the dynamic-attention stage produces $DAFF \in \mathbb{R}^{C \times H \times W}$ as shown in Eq.~\eqref{eq:daff}. This feature is used as the stabilized spatial-statistical input for the spectral refinement stage:
\begin{align}
\label{eq:daff}
DAFF &= \operatorname{Mona}\left(x + \operatorname{TSSA}\left(\operatorname{DyT}(x)\right)\right)
\end{align}

\begin{figure*}[t]
    \centering
    \includegraphics[
        width=\textwidth
    ]{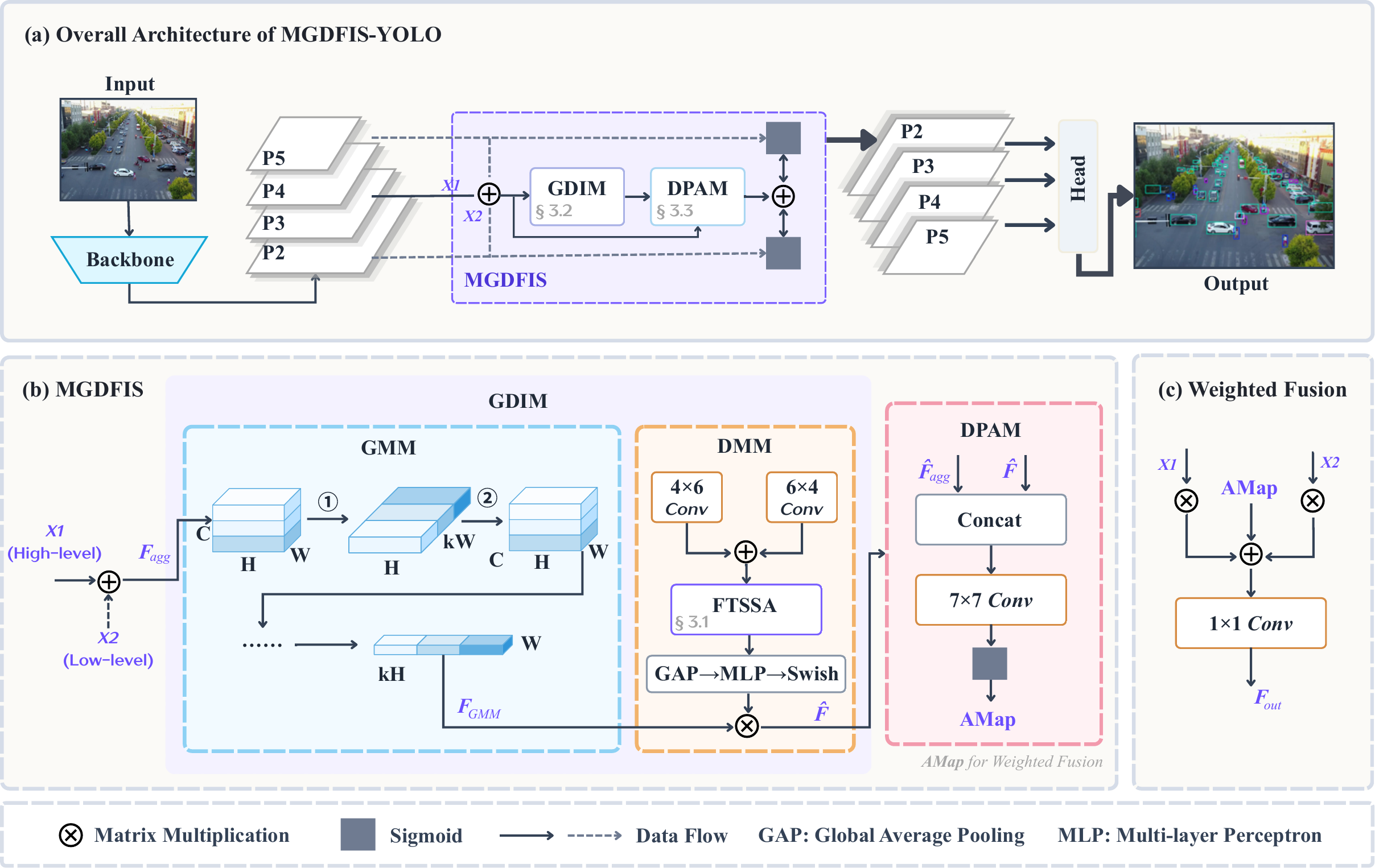}
    \caption{The \textbf{MGDFIS} framework connects global mixing, local-detail recovery, pixel-level recalibration, and weighted fusion for small-object feature fusion. (a) Overall MGDFIS-YOLO architecture. (b) Internal \textbf{MGDFIS} design, including \textbf{GDIM}, \textbf{GMM}, \textbf{DMM}, and \textbf{DPAM}. (c) Weighted fusion.}
    \label{fig:overview}
\end{figure*}

\subsubsection{Spectral Enhancement and Residual Refinement.}
The second stage complements the spatial attention path with frequency-domain refinement and residual stabilization. This design is useful because small UAV targets often appear as weak high-frequency structures whose boundaries can be diluted after repeated fusion. SEFF therefore reweights frequency responses before Mona Refinement restores local spatial structure through residual mixing.

In the frequency domain, we address both the instability of the spectral feature distribution and the lack of long-range dependency modeling in two steps. First, we reuse the DyT module, replacing the input feature $x$ in Eq.~\eqref{eq:dyt} with $DAFF$, to produce a spectrum-ready feature distribution $F_{S} \in \mathbb{R}^{C \times H \times W}$. Second, we adopt the Spectral Enhanced Feed-Forward (SEFF) block~\cite{sun2024hybrid}: the incoming distribution $F_{S}$ is split by a $1 \times 1$ convolution as shown in Eq.~\eqref{eq:seff_split},
\begin{align}
\label{eq:seff_split}
[F_{1},F_{2}] &= \operatorname{Conv}_{1 \times 1}(F_{S})
\end{align}
after which each branch extracts multi-scale features via depth-wise separable convolutions as shown in Eq.~\eqref{eq:seff_conv},
\begin{equation}
\label{eq:seff_conv}
\begin{aligned}
F_{1}' &= \operatorname{Conv}_{3\times3}^{d}(F_1), \\
F_{2}' &= \operatorname{Conv}_{3\times3}^{d,\mathrm{dil}}(F_2)
\end{aligned}
\end{equation}
where $\operatorname{Conv}_{3\times3}^{d,\mathrm{dil}}$ denotes a $3 \times 3$ dilated depth-wise separable convolution.

We then apply frequency-domain adaptive filtering to reweight these features, emphasizing useful spectral responses as shown in Eq.~\eqref{eq:frequency_filter}:
\begin{equation}
\label{eq:frequency_filter}
\begin{aligned}
F_{1}'' &= \mathcal{W}_1^{\uparrow} \odot \mathcal{F}(F_1')+\mathcal{B}_1, \\
F_{2}'' &= \mathcal{W}_2^{\uparrow} \odot \mathcal{F}(F_2')+\mathcal{B}_2
\end{aligned}
\end{equation}
where $\mathcal{W}_{i}^{\uparrow}\in \mathbb{C}^{C \times H \times W}$ and $\mathcal{B}_{i}\in \mathbb{R}^{C \times 1 \times 1}$ are the interpolated frequency-domain weight map and bias term for $i \in \{1,2\}$.

Following the gating mechanism, we combine these reweighted features with a SiLU activation to form gating signals, filtering the information as shown in Eq.~\eqref{eq:seff}: 
\begin{align}
\label{eq:seff}
F_{SEFF} &= \operatorname{Conv}_{1\times1}\!\left(\operatorname{SiLU}(\mathcal{F}^{-1}(F_2''))\odot\mathcal{F}^{-1}(F_1'')\right)
\end{align}
In the spatial domain, the SEFF-filtered feature $F_{SEFF} \in \mathbb{R}^{C \times H \times W}$ is then refined by Mona Refinement, which applies repeated multi-scale depth-wise separable convolutions with residual connections to reduce sensitivity to spatial distortions, as shown in Eq.~\eqref{eq:serr}: 
\begin{align}
\label{eq:serr}
SERR &= \operatorname{Mona}(DAFF + F_{SEFF}) 
\end{align}
Finally, Dynamic Attention and Feature Fusion and Spectral Enhancement and Residual Refinement are combined in series to form the FTSSA output. In the following module, this output is embedded into GDIM as the stabilized input for global-detail fusion.

\subsection{Global-Detail Integration Module}
\subsubsection{Spatial Concentration}

GMM starts from a compact observation: long-range interaction does not always require a heavy global operator if distant spatial positions are reorganized into a locally processable layout. This idea is consistent with grouped feature mixing, where feature rearrangement exposes non-local relations to lightweight convolutional operations. GMM applies this principle by slicing and reassembling feature maps along both row and column directions, allowing distant pixels to interact through compact convolutional neighborhoods, as shown in Fig.~\ref{fig:overview}(b).

For a given input feature $F \in \mathbb{R}^{C \times H \times W}$, GMM first partitions the channels into $k$ groups and rearranges them along the width direction, producing a column-concatenated feature $F_{Col} \in \mathbb{R}^{\frac{C}{k} \times H \times (kW)}$, where $k$ is the grouping factor. This transformation places originally separated positions into a shared local window, so a $3 \times 3$ convolution can exchange information beyond its original receptive field.

The rearranged feature is restored to $C$ channels by the $3 \times 3$ convolution, normalized by a Batch Normalization (BN) layer and a GELU activation to obtain $F_{restore}$, and then fused with the input feature $F$ through a $1 \times 1$ convolution to produce $F_{fuse} \in \mathbb{R}^{C \times H \times W}$.

Finally, the same operations are repeated along the row dimension: a 
$3 \times 3$ convolution and dimension reduction are performed, producing the final global mixed feature $F_{GMM} \in \mathbb{R}^{C \times H \times W}$. This abstraction keeps the operator easy to implement while giving local kernels access to globally rearranged context. The detailed pseudocode is provided in the supplementary material.

\subsubsection{Detail Capture}
After GMM reorganizes global context, the remaining issue is to recover directional details that are critical for compact objects. Visible-light images contain red, green, and blue (RGB) channels, but their spatial layout carries edges, elongated shapes, and local texture changes that are important for tiny vehicles. DMM is therefore designed as a local-detail recovery module: it uses directional convolutions and attention weighting to strengthen anisotropic spatial cues after global mixing, as shown in Fig.~\ref{fig:overview}(b).

For the feature $F_{GMM} \in \mathbb{R}^{C \times H \times W}$ passed from GMM, we use convolution kernels of different orientations to extract complementary spatial evidence. Padding and channel projection are applied inside the directional convolutions so that the extracted local features $F_{D_1}$ and $F_{D_2}$ have the same shape as $F_{GMM}$. These features are then combined with $F_{GMM}$ to obtain the fused feature $F_{Add}\in \mathbb{R}^{C \times H \times W}$, which contains both global and local information:
\begin{equation}
\label{eq:dmm_dir}
\begin{aligned}
  F_{D_1} &= \operatorname{Conv}_{4\times6}(F_{GMM}), \\
  F_{D_2} &= \operatorname{Conv}_{6\times4}(F_{GMM})
\end{aligned}
\end{equation}
\begin{equation}
\label{eq:dmm_add}
  F_{Add} = F_{GMM} + F_{D_1} + F_{D_2}
\end{equation}

To refine the fused feature $F_{Add}$, we first apply FTSSA to filter its spectral-spatial responses. We then perform global average pooling (GAP), use a Multilayer Perceptron (MLP) for feature compression, and apply Swish to produce the channel attention:
\begin{equation}
\label{eq:dmm_att}
\operatorname{Att}(F_{Add})=\operatorname{Swish}(\operatorname{MLP}(\operatorname{GAP}(\operatorname{FTSSA}(F_{Add}))))
\end{equation}
where $\operatorname{Swish}(x)=x \cdot \sigma(x)$ and $\sigma(\cdot)$ denotes the sigmoid function.
The final output of DMM is defined in Eq.~\eqref{eq:dmm}:
\begin{equation}
\label{eq:dmm}
\operatorname{DMM}(F_{GMM})=F_{Add} \odot \operatorname{Att}(F_{Add})
\end{equation}

\begin{table*}[t]
\centering
\caption{Comparison with recent detectors on VisDrone. \textcolor{bestred}{Best}/\textcolor{secblue}{second-best} per accuracy column colored; shaded row is ours. $AP_{L}$ is dropped, as VisDrone has almost no large instances.}
\label{tab:visdrone_pdf_models}
\scriptsize
\setlength{\tabcolsep}{3pt}
\renewcommand{\arraystretch}{1.1}
\begin{tabularx}{\textwidth}{ll*{8}{>{\centering\arraybackslash}X}}
\toprule
Model & Venue & Params & GFLOPs & FPS & $AP_{50}$ & $AP_{75}$ & $AP_{50:95}$ & $AP_{S}$ & $AP_{M}$ \\
\midrule
\multicolumn{10}{l}{\textit{Recent detectors}}\\
RT-DETR-R18~\cite{zhao2024detrs}        & CVPR'24    & 20.0 & 60.0  & 76.0 & 45.8 & 28.5 & 27.7 & 18.5 & 39.5 \\
Drone-DETR~\cite{kong2024drone}         & Sensors'24 & 28.7 & 128.3 & 30.0 & 42.4 & 25.1 & 24.9 & 17.8 & 35.3 \\
DINO~\cite{zhang2023dino}               & ICLR'23    & 47.0 & 279.0 & 24.0 & 46.2 & 23.6 & 24.9 & 16.9 & 35.6 \\
Deformable DETR~\cite{zhu2021deformable}& ICLR'21    & 10.0 & 196.0 & 29.0 & 42.2 & 28.9 & 27.1 & 15.6 & 39.6 \\
CFPT~\cite{du2025cross}                 & TGRS'25    & 56.3 & 297.6 & --   & 38.0 & 23.1 & 22.6 & 12.1 & 36.2 \\
QueryDet~\cite{yang2022querydet}        & CVPR'22    & 33.9 & 212.0 & --   & 48.1 & 28.8 & 28.3 & 19.6 & 39.8 \\
UAV-DETR-R18~\cite{zhang2025uav}        & arXiv'25   & 20.0 & 77.0  & --   & \textcolor{secblue}{48.8} & 30.8 & 29.8 & 20.7 & 41.7 \\
UAV-DETR-R50~\cite{zhang2025uav}        & arXiv'25   & 42.0 & 170.0 & --   & \textcolor{bestred}{51.1} & \textcolor{secblue}{32.8} & \textcolor{secblue}{31.5} & \textcolor{secblue}{22.5} & \textcolor{secblue}{43.4} \\
\midrule
\multicolumn{10}{l}{\textit{YOLO-P2 baselines}}\\
YOLO11m-P2~\cite{jocher2024yolo11}      & Ultralytics'24 & 20.7 & 91.3 & 72.1 & 46.4 & 27.6 & 27.3 & 18.4 & 38.8 \\
YOLOv12m-P2~\cite{tian2025yolov12}      & arXiv'25       & 20.0 & 77.7 & 74.2 & 36.2 & 20.9 & 21.0 & 11.5 & 32.8 \\
YOLO26m-P2~\cite{ultralytics2025yolo26} & Ultralytics'25 & 21.1 & 91.4 & 69.2 & 42.1 & 29.4 & 27.4 & 17.3 & 40.0 \\
\midrule
\rowcolor{mgshade}
\textbf{YOLO26m-P2+MGDFIS} & \textbf{Ours} & \textbf{28.2} & \textbf{119.6} & \textbf{67.3} & \textbf{47.8} & \textcolor{bestred}{\textbf{36.5}} & \textcolor{bestred}{\textbf{34.1}} & \textcolor{bestred}{\textbf{23.1}} & \textcolor{bestred}{\textbf{47.7}} \\
\bottomrule
\end{tabularx}
\end{table*}

\subsubsection{Global-Detail Fusion}

GDIM arranges GMM and DMM in series so that global context exchange precedes local-detail recovery, as shown in Fig.~\ref{fig:overview}(b). The original features $F_{O_1}\in \mathbb{R}^{C_1 \times H_1 \times W_1}$ and $F_{O_2}\in \mathbb{R}^{C_2 \times H_2 \times W_2}$ are first aligned to a common channel width and spatial resolution. Specifically, each alignment operator $\mathcal{A}_i(\cdot)$ contains a $1 \times 1$ channel projection and a scale adjustment operation, bilinear interpolation for higher resolution or strided convolution for lower resolution. The symbol $+$ denotes element-wise summation after this alignment, rather than direct addition of mismatched tensors. After passing $F_{agg}$ through GMM, it is transformed into the global mixed feature $F_{GMM}$, which is then fed into DMM for local-detail refinement, producing the global-detail complementary enhanced feature $\hat{F}\in \mathbb{R}^{C \times H \times W}$. The computation is defined in Eqs.~\eqref{eq:gdim_align}-\eqref{eq:gdim_out}:
\begin{align}
  \tilde{F}_{O_i} &= \mathcal{A}_i(F_{O_i}) \in \mathbb{R}^{C \times H \times W},\quad i\in\{1,2\}
  \label{eq:gdim_align}\\
  F_{agg} &= \tilde{F}_{O_1} + \tilde{F}_{O_2}
  \label{eq:gdim_agg}
\end{align}
\vspace{-6pt}
\begin{equation}
  \label{eq:gdim_out}
  \hat{F} = \operatorname{DMM}\bigl(\operatorname{GMM}(F_{agg})\bigr)
\end{equation}
\vspace{-3pt}
\subsection{Dynamic Pixel Attention Module}

DPAM is introduced to convert the global-detail representation into pixel-level foreground selection. Even after GDIM, cluttered aerial backgrounds can still activate around roads, roofs, and repeated textures. The DPAM structure is shown in Fig.~\ref{fig:overview}(b). The aggregated feature $F_{agg}$ and the refined feature $\hat{F}$, obtained from the input and GDIM respectively, are fed into DPAM to generate an attention weight map $\mathrm{AMap}\in [0,1]^{C\times H\times W}$, which encodes both the coarse-detail information from $F_{agg}$ and the fine-grained information from $\hat{F}$.

Specifically, given $F_{agg}$ and $\hat{F}$, we first concatenate them along the channel dimension to form a mixed feature map:
\begin{equation}
\label{eq:dpam_mix}
F_{Mix} = \operatorname{Concat}(F_{agg},\hat{F})
\end{equation}
Next, we apply a spatial convolution to $F_{Mix}$ to obtain the refined local feature $F_{Loc}\in \mathbb{R}^{C\times H\times W}$:
\begin{equation}
\label{eq:dpam_loc}
F_{Loc}=\operatorname{Conv}_{7 \times 7}(F_{Mix})
\end{equation}
Finally, we normalize $F_{Loc}$ with a sigmoid activation to produce the pixel-wise attention weight map, as defined in Eq.~\eqref{eq:dpam}:
\begin{equation}
\label{eq:dpam}
\operatorname{DPAM}(F_{agg},\hat{F})=\mathrm{AMap}=\sigma (F_{Loc})
\end{equation}

\subsection{MGDFIS as a Feature Fusion Strategy}
After DPAM produces the attention map, the aligned features are recalibrated at the pixel level. The final fusion combines the DPAM-weighted representation with the aligned source features, so base information and refined foreground cues remain in the same path. MGDFIS therefore organizes the full path as stabilization, global-detail integration, and pixel-level recalibration, addressing receptive-field mismatch while keeping the feature flow traceable from input to output.
\vspace{-2pt}
\section{Experiments and Results}
\subsection{Experimental Setup}
\noindent\textbf{Datasets.}
We evaluate across three aerial regimes that stress different parts of the small-object problem. VisDrone-2019-DET is our primary, controlled benchmark: dense static scenes with severe scale variation, occlusion, and background clutter. UAVDT shifts the stressor to motion and viewpoint change in UAV video, and CARPK to the opposite extreme of regular, well-separated cars where coarse detection is nearly saturated and only localization precision remains to be won.

\noindent\textbf{Metrics.}
$AP_{50:95}$ is our metric of record and $AP_{S}$ is the primary small-object metric in our evaluation, since it measures accuracy on the objects MGDFIS is built for; we report $AP_{50}$, $AP_{75}$, $AP_{M}$, parameters, GFLOPs, and FPS alongside them.

\noindent\textbf{Implementation.}
MGDFIS is instantiated on YOLO26m and trained with AdamW for 400 epochs (input $640$, batch $64$, lr $0.001$, momentum $0.937$, weight decay $0.0005$) on a single NVIDIA A800. All paired variants follow the same dataset-specific protocol and differ only by the neck; strict organization-only attribution requires a FLOP-matched YOLO26m default-neck control, with further details in the supplementary material. 


\subsection{Main Results on VisDrone and UAVDT}
MGDFIS improves every YOLO generation it is attached to. On YOLO26m it adds $4.5$ $AP_{50:95}$ ($25.7\!\to\!30.2$) and $7.0$ $AP_{50}$ ($37.2\!\to\!44.2$); with the P2 head the two effects compound to $34.1/47.8$.

The gains are stable across four detector generations. From YOLOv8 through YOLO26, the same neck adds between $5$ and $10$ $AP_{50}$ to both the m and m-P2 variants, with an essentially \emph{constant} $+28.2$ GFLOPs and $\sim\!4.6$M parameters in every family. These paired rows show that MGDFIS is an effective neck replacement; the capacity-control protocol above is required for organization-only attribution. We discuss the attribution boundary of this neck-replacement comparison in the supplementary material. Full generation-wise results are provided in the supplementary material.

\begin{figure*}[t]
    \centering
    \includegraphics[width=\linewidth]{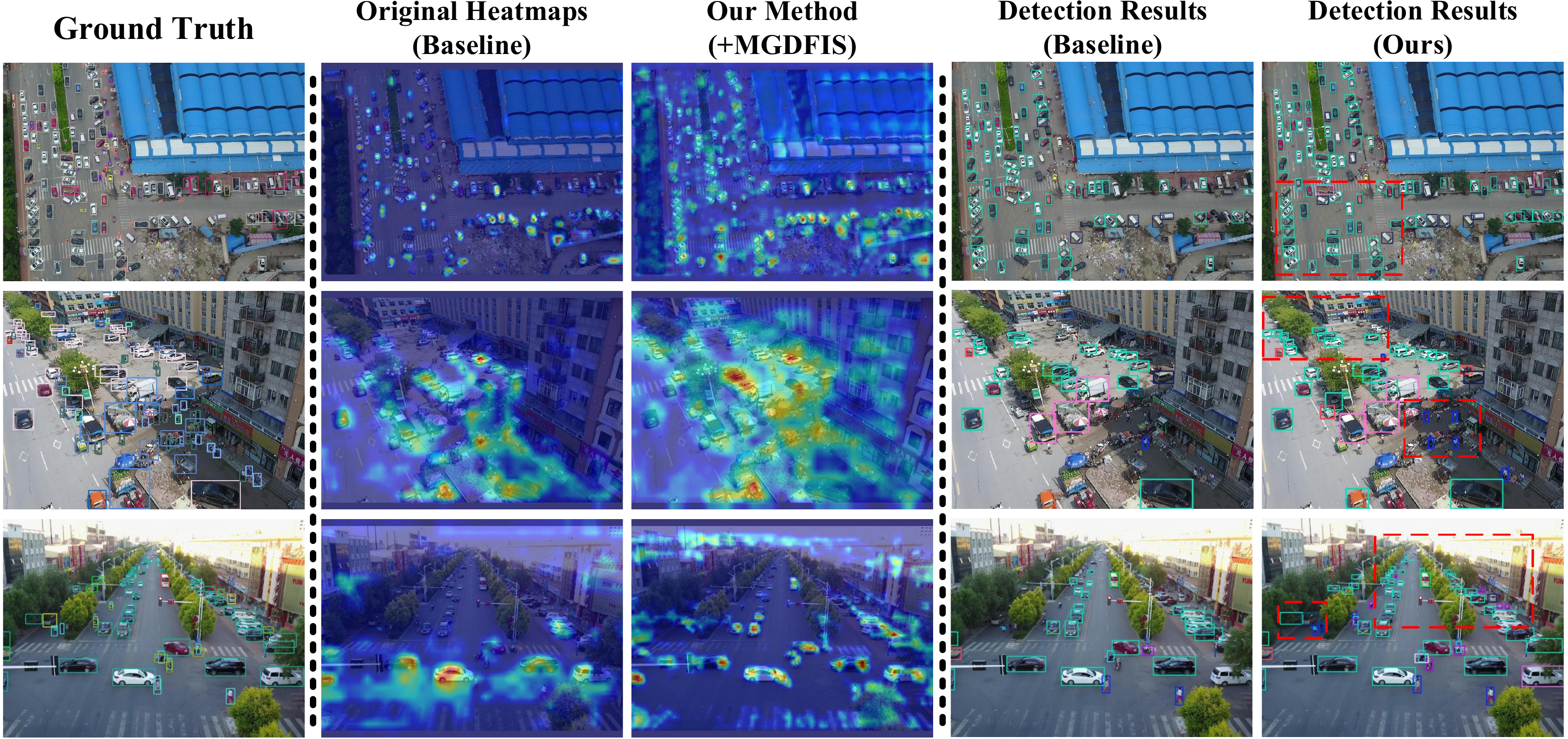}
    \caption{\textbf{VisDrone heatmap comparison.} Compared with the baseline, MGDFIS concentrates activation on compact vehicle regions and suppresses responses on road markings, rooftops, and other cluttered background areas.}
    \label{fig:heatmap}
    \vspace{-0.5em}
\end{figure*}

Tables~\ref{tab:visdrone_pdf_models} and~\ref{tab:yolo_ablation}, together with Fig.~\ref{fig:heatmap}, summarize the main comparison, component contribution, and qualitative evidence before the detailed discussion below.

\begin{table}[!tb]
\centering
\caption{Cumulative ablation on VisDrone (YOLO26m). $\Delta$: per-module change in $AP_{50:95}$ over the row above; the all-``--'' row is the baseline. \textbf{Bold}: best $AP$ per block; shaded: full model.}
\label{tab:yolo_ablation}
\scriptsize
\setlength{\tabcolsep}{3.2pt}
\renewcommand{\arraystretch}{1.2}
\begin{tabularx}{\columnwidth}{cccc*{4}{>{\centering\arraybackslash}X}}
\toprule
GMM & DMM & FTSSA & DPAM & $AP_{50:95}$ & $\Delta$ & $AP_{50}$ & GFLOPs \\
\midrule
\multicolumn{8}{l}{\textit{With P2 head}}\\
--         & --         & --         & --         & 27.4 & --     & 42.1 & 91.4 \\
\checkmark & --         & --         & --         & 28.1 & $+0.7$ & 43.2 & 105.1 \\
\checkmark & \checkmark & --         & --         & 30.7 & $+2.6$ & 44.3 & 106.2 \\
\checkmark & \checkmark & \checkmark & --         & 33.1 & $+2.4$ & 45.7 & 117.2 \\
\rowcolor{mgshade}
\checkmark & \checkmark & \checkmark & \checkmark & \textbf{34.1} & $+1.0$ & \textbf{47.8} & 119.6 \\
\bottomrule
\end{tabularx}
\end{table}

On UAVDT the neck transfers without retuning (Table~\ref{tab:uavdt_pdf_models}). It lifts YOLO26m-P2 by $+6.9$ $AP_{50}$ and $+6.1$ $AP_{50:95}$, and adds a comparable $5$--$6$ points to the other three families. Under the reported metrics, all four MGDFIS rows outperform the listed published baselines, including the cluster- and context-guided detectors (ClusDet, GLSAN, CEASC) and the transformer detectors (RT-DETR, Deformable DETR). This result suggests that preserving weak vehicle evidence during fusion is a key bottleneck on UAVDT.

\begin{table}[!tb]
\centering
\caption{UAVDT comparison. \textcolor{bestred}{Best}/\textcolor{secblue}{second-best} per accuracy column colored.}
\label{tab:uavdt_pdf_models}
\scriptsize
\setlength{\tabcolsep}{2.6pt}
\renewcommand{\arraystretch}{1.12}
\begin{tabularx}{\linewidth}{@{}>{\raggedright\arraybackslash}Xrrrr@{}}
\toprule
Method & GFLOPs & $AP_{50}$ & $AP_{75}$ & $AP_{50:95}$ \\
\midrule
\multicolumn{5}{@{}l}{\textit{Published baselines}}\\
ClusDet~\cite{yang2019clustered}         & --    & 26.5 & 12.5 & 13.7 \\
GLSAN~\cite{deng2021glsan}               & --    & 28.1 & 18.8 & 17.0 \\
DREN~\cite{zhang2019dren}                & --    & --   & --   & 15.1 \\
GFL~\cite{li2020generalized}             & --    & 29.5 & 17.9 & 16.9 \\
CEASC~\cite{du2023adaptive}              & --    & 30.9 & 17.8 & 17.1 \\
RT-DETR-R18~\cite{zhao2024detrs}         & 60.0  & 33.2 & 19.2 & 16.5 \\
Deformable DETR~\cite{zhu2021deformable} & 196.0 & 32.7 & 16.9 & 15.9 \\
\midrule
\multicolumn{5}{@{}l}{\textit{YOLO-P2 references}}\\
YOLOv8m-P2~\cite{jocher2023yolov8}       & 98.0  & 31.4 & 17.2 & 16.9 \\
YOLO11m-P2~\cite{jocher2024yolo11}       & 91.3  & 32.6 & 17.6 & 15.4 \\
YOLOv12m-P2~\cite{tian2025yolov12}       & 77.7  & 31.9 & 16.9 & 14.5 \\
YOLO26m-P2~\cite{ultralytics2025yolo26}  & 91.4  & 32.0 & 17.4 & 15.8 \\
\midrule
\multicolumn{5}{@{}l}{\textit{Ours\,(\,$+$\,MGDFIS\,)}}\\
\rowcolor{mgshade} YOLOv8m-P2 + MGDFIS   & 126.2 & 36.8 & 21.5 & 20.1 \\
\rowcolor{mgshade} YOLO11m-P2 + MGDFIS   & 119.5 & \textcolor{secblue}{38.4} & \textcolor{secblue}{22.8} & \textcolor{secblue}{21.6} \\
\rowcolor{mgshade} YOLOv12m-P2 + MGDFIS  & 105.9 & 37.1 & 21.8 & 20.4 \\
\rowcolor{mgshade} \textbf{YOLO26m-P2 + MGDFIS} & \textbf{119.6} & \textcolor{bestred}{\textbf{38.9}} & \textcolor{bestred}{\textbf{23.1}} & \textcolor{bestred}{\textbf{21.9}} \\
\bottomrule
\end{tabularx}
\end{table}

\subsection{Comparison with Recent Detectors}
Read column by column, Table~\ref{tab:visdrone_pdf_models} shows a clear signature rather than a uniform win. MGDFIS holds the best $AP_{75}$, $AP_{50:95}$, $AP_{S}$ and $AP_{M}$, but ranks third on $AP_{50}$ ($47.8$, behind UAV-DETR's $51.1$ and $48.8$). This pattern indicates that the neck mainly improves \emph{accurate, tightly localized} detections of small objects (high $AP_{S}$, high-IoU $AP$), rather than coarse recall (loose $AP_{50}$). The compute follows the same pattern. MGDFIS reaches $34.1$ $AP_{50:95}$ at $119.6$ GFLOPs, while the methods it overtakes on strict accuracy, DINO ($279$), CFPT ($297.6$), and Deformable DETR ($196$), spend $1.6$ to $2.5\times$ more. The result places MGDFIS in the high-strict-accuracy, moderate-compute corner of the trade-off; the matched-cost row is the comparison for capacity-controlled attribution.

Across all three datasets MGDFIS adds five to seven points of $AP_{50:95}$ to every P2 detector for a near-constant compute increment; the improvement and its price are both stable properties of the neck replacement rather than dataset- or detector-specific accidents.

\subsection{CARPK Evaluation}
CARPK tests the regime where coarse detection is already solved: $AP_{50}$ saturates near $95$ for all baselines, and every method clusters within a single point, with MGDFIS edging to the top at $97.4$ (Table~\ref{tab:carpk_pdf_models}). The discriminative column is $AP_{50:95}$, where MGDFIS tightens the YOLO baselines from $\sim\!58$ to $62.3$, corresponding to a four-point localization gain after coarse detection stops being the bottleneck. Drone-YOLO-M reports a higher $68.3$ from a heavier specialized pipeline; our claim is narrower: the same neck that helps under VisDrone clutter also tightens localization on clean parking layouts, which suggests that its benefit is related to localization precision rather than a specific background type.

\begin{table}[!tb]
\centering
\caption{CARPK comparison. \textbf{Bold}: best among the YOLO rows; shaded rows are ours.}
\label{tab:carpk_pdf_models}
\scriptsize
\setlength{\tabcolsep}{6pt}
\renewcommand{\arraystretch}{1.18}
\begin{tabularx}{\linewidth}{l*{3}{>{\centering\arraybackslash}X}}
\toprule
Method & GFLOPs & $AP_{50:95}$ & $AP_{50}$ \\
\midrule
\multicolumn{4}{l}{\textit{Published references}}\\
DINO~\cite{zhang2023dino}               & 279.0 & 49.6 & 94.4 \\
Drone-YOLO-M~\cite{zhang2023drone}      & --    & 68.3 & 96.6 \\
\midrule
\multicolumn{4}{l}{\textit{YOLO-P2 baselines}}\\
YOLOv8m-P2~\cite{jocher2023yolov8}      & 98.0  & 57.4 & 95.1 \\
YOLO11m-P2~\cite{jocher2024yolo11}      & 91.3  & 58.2 & 95.3 \\
YOLOv12m-P2~\cite{tian2025yolov12}      & 77.7  & 58.0 & 95.2 \\
YOLO26m-P2~\cite{ultralytics2025yolo26} & 91.4  & 58.1 & 95.4 \\
\midrule
\multicolumn{4}{l}{\textit{Ours\,(\,$+$\,MGDFIS\,)}}\\
\rowcolor{mgshade} YOLOv8m + MGDFIS   & 107.1 & 60.2 & 96.8 \\
\rowcolor{mgshade} YOLOv12m + MGDFIS  & 95.7  & 61.5 & 97.1 \\
\rowcolor{mgshade} YOLO26m + MGDFIS   & 96.1  & \textbf{62.3} & \textbf{97.4} \\
\bottomrule
\end{tabularx}
\end{table}


\subsection{Qualitative Analysis}
The heatmaps in Fig.~\ref{fig:heatmap} make the mechanism visible. On tiny and partially occluded vehicles the baseline leaks activation onto road markings and rooftops, whereas MGDFIS contracts the response onto the compact foreground. This is the spatial counterpart of DPAM's pixel-level recalibration and the direct cause of the higher $AP_{S}$ in the tables.

\subsection{Ablation Study}
The ablation (Table~\ref{tab:yolo_ablation}) traces the gain to its sources under the final P2 setting. GMM provides the scaffold for long-range mixing, DMM is the largest single lever ($+2.6$), and FTSSA contributes a further $+2.4$ by stabilizing weak responses before they are fused. DPAM then adds $+1.0$ $AP_{50:95}$ and raises $AP_{50}$ to $47.8$, showing that pixel-level foreground reweighting becomes effective when high-resolution features provide a precise map to act on. Additional without-P2 ablation in the supplementary material shows the complementary case, where DPAM improves coarse recall but can hurt strict localization on lower-resolution maps.

\vspace{-6pt}
\section{Conclusion}
We have presented MGDFIS for UAV small-object detection, a neck-level feature-fusion strategy that links feature stabilization, global-detail integration, and pixel-level foreground-background recalibration. The key idea is to preserve weak local evidence while improving cross-scale context exchange. On VisDrone, adding MGDFIS to YOLO26m improves $AP_{50:95}$ from 25.7 to 30.2 and $AP_{50}$ from 37.2 to 44.2; with a P2 head the full configuration reaches 34.1 $AP_{50:95}$ and 47.8 $AP_{50}$. The gain transfers across four YOLO generations and across UAVDT and CARPK at a consistent neck cost, while a capacity-matched default-neck control is required for organization-only attribution.
The main cost is computation: the neck adds a fixed ${\sim}28$ GFLOPs, raising YOLO26m from 67.9 to 96.1 (119.6 with the P2 head). Future work should study lighter fusion variants and deployment-oriented efficiency.
{
    \small
    \bibliographystyle{ieeenat_fullname}
    \bibliography{references}
}

\clearpage
\appendix
\twocolumn[
\begin{@twocolumnfalse}
\begin{center}
    {\LARGE \textbf{Technical Appendices and Supplementary Material}}
\end{center}
\vspace{1em}
\end{@twocolumnfalse}
]
\begin{figure*}[t]
    \centering
    \includegraphics[
        width=\textwidth,
        trim=50pt 8pt 10pt 8pt,
        clip
    ]{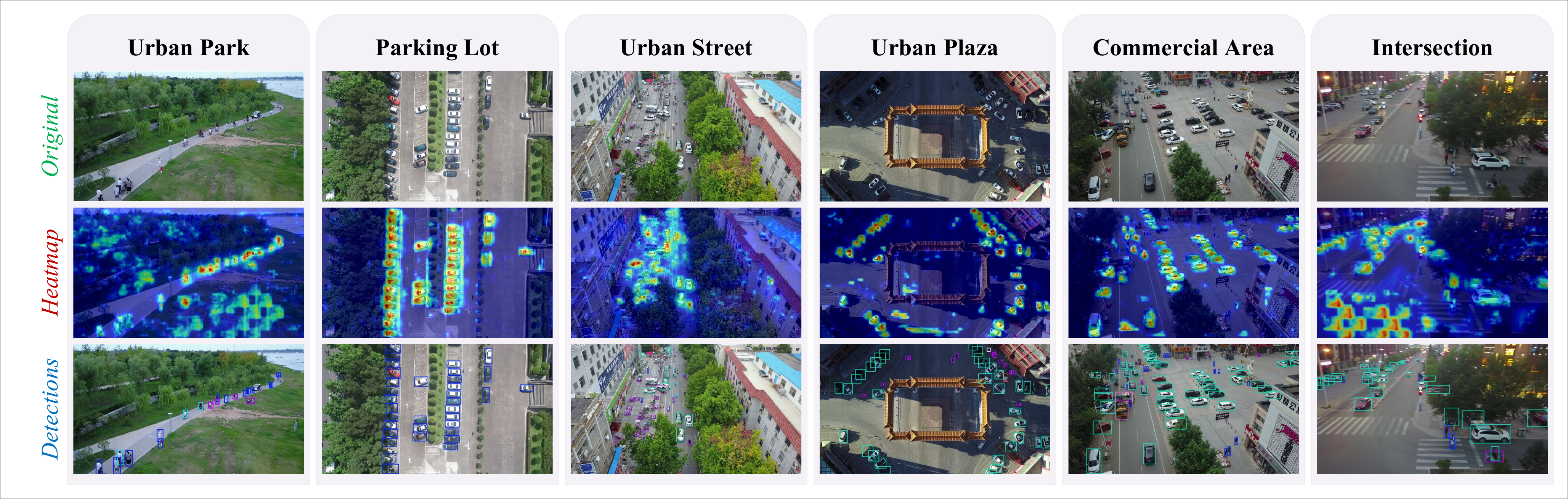}
    \caption{
        Supplementary VisDrone illustration. Dense, dispersed, and weakly resolved targets motivate the multi-scale feature integration in MGDFIS.
    }
    \label{fig:visdrone_supp}
\end{figure*}

\section{Additional Details}
\vspace{-0.35em}

This appendix provides implementation and experimental details that support the main paper. Figure~\ref{fig:visdrone_supp} illustrates the VisDrone small-object setting, where targets are dense, spatially dispersed, and weakly resolved. Sec.~A.1 gives the step-by-step pseudocode of the Global Mixing Module (GMM); Sec.~A.2 tests MGDFIS across YOLOv8, YOLO11, YOLOv12, and YOLO26; Sec.~A.3 isolates the non-P2 YOLO26m module sequence; and Sec.~A.4 clarifies how to interpret the paired comparison in Table~\ref{tab:yolo_generation_supp}.

\subsection{GMM Pseudocode}
\vspace{-0.25em}

This section details the Global Mixing Module (GMM) introduced in Sec.~3.2.1 of the main paper. GMM enlarges the effective spatial context by concentrating features along the width and height directions. In each direction, it splits the feature map into channel groups, concatenates the groups along one spatial axis, applies a lightweight local convolution, restores the original layout, and fuses the restored feature with the input representation. This exposes distant positions to a local convolutional neighborhood without using a heavy global operator. For consistency with Sec.~3.2.3, Algorithm~\ref{alg:gmm} also includes the GDIM alignment and summation step before the width-wise and height-wise GMM operations begin.

\begin{algorithm}[t]
\caption{Global Mixing Module (GMM)}
\label{alg:gmm}
\small
\begin{algorithmic}[1]
\Require Features $F_{O_1} \in \mathbb{R}^{C_1 \times H_1 \times W_1}$,
         $F_{O_2} \in \mathbb{R}^{C_2 \times H_2 \times W_2}$; grouping factor $k$
\Ensure Global mixed feature $F_{GMM} \in \mathbb{R}^{C \times H \times W}$

\State $\tilde{F}_{O_1} \gets \mathcal{A}_1(F_{O_1})$, 
       $\tilde{F}_{O_2} \gets \mathcal{A}_2(F_{O_2})$
\State $F \gets \tilde{F}_{O_1} + \tilde{F}_{O_2}$

\Statex \textit{// Width-wise mixing}
\State Split $F$ into $k$ channel groups $\{F_j\}_{j=1}^{k}$
\State $F^{w}_{cat} \gets \mathrm{Concat}_{w}(F_1,\ldots,F_k)$
\State $F^{w}_{mix} \gets \mathrm{GELU}(\mathrm{BN}(\mathrm{Conv}_{3\times3}(F^{w}_{cat})))$
\State $F^{w}_{res} \gets \mathrm{Restore}_{w}(F^{w}_{mix})$
\State $F \gets \mathrm{Conv}_{1\times1}(\mathrm{Concat}(F, F^{w}_{res}))$

\Statex \textit{// Height-wise mixing}
\State Split $F$ into $k$ channel groups $\{F_j\}_{j=1}^{k}$
\State $F^{h}_{cat} \gets \mathrm{Concat}_{h}(F_1,\ldots,F_k)$
\State $F^{h}_{mix} \gets \mathrm{GELU}(\mathrm{BN}(\mathrm{Conv}_{3\times3}(F^{h}_{cat})))$
\State $F^{h}_{res} \gets \mathrm{Restore}_{h}(F^{h}_{mix})$
\State $F \gets \mathrm{Conv}_{1\times1}(\mathrm{Concat}(F, F^{h}_{res}))$

\State $F_{GMM} \gets F$
\State \Return $F_{GMM}$
\end{algorithmic}
\end{algorithm}

The restore operations invert the spatial concentration in the two mixing passes. $\mathrm{Restore}_{w}(\cdot)$ splits the width-concentrated feature into $k$ chunks along the width axis and concatenates them back along the channel axis, and $\mathrm{Restore}_{h}(\cdot)$ applies the same operation to the height-concentrated feature. The two passes are executed sequentially, so column-wise and row-wise long-range interactions are incorporated before DMM performs local-detail refinement.

\subsection{Generation-wise YOLO Results}
\vspace{-0.25em}

Table~\ref{tab:yolo_generation_supp} reports generation-wise results on VisDrone. Each MGDFIS row is paired with the corresponding YOLO baseline under the same training protocol, so the table measures the effect of replacing the neck while keeping the detector family fixed. MGDFIS improves both the standard $m$ variants and the $m$-P2 variants across YOLOv8, YOLO11, YOLOv12, and YOLO26, with $AP_{50}$ gains of $+4.7$ to $+10.5$ and $AP_{50:95}$ gains of $+4.5$ to $+8.6$. These paired comparisons show that the benefit is not tied to a single YOLO generation; the next subsection isolates the non-P2 module sequence behind the YOLO26m comparison.

\begin{table}[H]
\centering
\caption{YOLO-family comparison on VisDrone. Each $+\mathrm{MGDFIS}$ row is paired with the corresponding baseline, and the $\Delta$ columns report the absolute gain over the row above. The YOLO26m rows connect this comparison to the non-P2 ablation in Table~\ref{tab:ablation_without_p2_supp}.}
\label{tab:yolo_generation_supp}
\scriptsize
\setlength{\tabcolsep}{2.0pt}
\renewcommand{\arraystretch}{1.02}
\resizebox{\columnwidth}{!}{%
\begin{tabular}{lrrrrrr}
\toprule
\textbf{Model} & \textbf{$AP_{50}$} & \textbf{$\Delta_{50}$} & \textbf{$AP_{50:95}$} & \textbf{$\Delta_{50:95}$} & \textbf{Par.} & \textbf{FLOPs} \\
\midrule
\rowcolor{yologroupbg}
\multicolumn{7}{l}{\emph{YOLOv8~\cite{yolov8}}} \\
YOLOv8m & 40.7 & -- & 24.6 & -- & 25.9 & 78.9 \\
YOLOv8m + MGDFIS & 49.7 & +9.0 & 33.2 & +8.6 & 30.2 & 107.1 \\
YOLOv8m-P2 & 49.2 & -- & 30.3 & -- & 25.0 & 98.0 \\
YOLOv8m-P2 + MGDFIS & 53.9 & +4.7 & 35.2 & +4.9 & 33.2 & 126.2 \\
\midrule
\rowcolor{yologroupbg}
\multicolumn{7}{l}{\emph{YOLO11~\cite{yolo11}}} \\
YOLO11m & 43.3 & -- & 26.3 & -- & 20.1 & 68.2 \\
YOLO11m + MGDFIS & 48.1 & +4.8 & 32.1 & +5.8 & 25.1 & 96.4 \\
YOLO11m-P2 & 46.4 & -- & 27.3 & -- & 20.7 & 91.3 \\
YOLO11m-P2 + MGDFIS & 52.3 & +5.9 & 34.5 & +7.2 & 27.3 & 119.5 \\
\midrule
\rowcolor{yologroupbg}
\multicolumn{7}{l}{\emph{YOLOv12~\cite{yolov12}}} \\
YOLOv12m & 33.6 & -- & 19.2 & -- & 20.2 & 67.5 \\
YOLOv12m + MGDFIS & 42.1 & +8.5 & 25.8 & +6.6 & 24.3 & 95.7 \\
YOLOv12m-P2 & 36.2 & -- & 21.0 & -- & 20.0 & 77.7 \\
YOLOv12m-P2 + MGDFIS & 46.7 & +10.5 & 28.9 & +7.9 & 26.3 & 105.9 \\
\midrule
\rowcolor{yologroupbg}
\multicolumn{7}{l}{\emph{YOLO26~\cite{yolo26}}} \\
YOLO26m & 37.2 & -- & 25.7 & -- & 20.4 & 67.9 \\
YOLO26m + MGDFIS & 44.2 & +7.0 & 30.2 & +4.5 & 25.0 & 96.1 \\
YOLO26m-P2 & 42.1 & -- & 27.4 & -- & 21.1 & 91.4 \\
YOLO26m-P2 + MGDFIS & 47.8 & +5.7 & 34.1 & +6.7 & 28.2 & 119.6 \\
\bottomrule
\end{tabular}%
}
\end{table}

\vspace{-0.35em}

\subsection{Ablation Without P2 Head}
\vspace{-0.25em}

Table~\ref{tab:ablation_without_p2_supp} complements the YOLO26m rows in Table~\ref{tab:yolo_generation_supp} with a cumulative ablation without the P2 head. The first three components mainly improve strict localization: GMM provides long-range spatial concentration, DMM recovers directional details, and FTSSA stabilizes spectral-spatial responses before fusion. DPAM raises $AP_{50}$ from $42.2$ to $44.2$, but reduces $AP_{50:95}$ from $31.8$ to $30.2$. This suggests that pixel-level foreground reweighting helps coarse foreground activation at lower resolution, whereas strict localization still depends on the finer spatial evidence provided by the P2 head.

\begin{table}[H]
\centering
\caption{Cumulative ablation on VisDrone without the P2 head. $\Delta$ reports the per-module change in $AP_{50:95}$ over the row above, and the all-disabled row is the baseline.}
\label{tab:ablation_without_p2_supp}
\scriptsize
\setlength{\tabcolsep}{3.0pt}
\renewcommand{\arraystretch}{1.03}
\resizebox{\columnwidth}{!}{%
\begin{tabular}{ccccrrrr}
\toprule
\textbf{GMM} & \textbf{DMM} & \textbf{FTSSA} & \textbf{DPAM} & \textbf{$AP_{50:95}$} & \textbf{$\Delta$} & \textbf{$AP_{50}$} & \textbf{GFLOPs} \\
\midrule
-- & -- & -- & -- & 25.7 & -- & 37.2 & 67.9 \\
\cmark & -- & -- & -- & 26.3 & +0.6 & 38.9 & 81.6 \\
\cmark & \cmark & -- & -- & 29.4 & +3.1 & 39.8 & 81.9 \\
\cmark & \cmark & \cmark & -- & 31.8 & +2.4 & 42.2 & 95.5 \\
\cmark & \cmark & \cmark & \cmark & 30.2 & -1.6 & 44.2 & 96.1 \\
\bottomrule
\end{tabular}%
}
\end{table}
\vspace{-0.35em}

\subsection{Discussion on Capacity and Attribution}
\vspace{-0.25em}

The paired rows in Table~\ref{tab:yolo_generation_supp} measure the practical effect of replacing the YOLO neck with MGDFIS under the same training protocol. This comparison should not be interpreted as pure organization-only attribution, because MGDFIS also changes the parameter count and computational budget. A stricter attribution study would keep the default YOLO neck design, widen or deepen it until its GFLOPs match MGDFIS, and train it under the identical protocol. Such a control would separate gains from feature-organization design from gains caused by extra capacity. We therefore treat the results as evidence for MGDFIS as an effective neck replacement, while leaving capacity-matched default-neck controls to future work.

\end{document}